\documentclass[5p]{elsarticle}

\usepackage{algorithm}
\usepackage{algpseudocode}
\usepackage{amsmath}
\usepackage{amssymb}
\usepackage{color}
\usepackage{graphicx}
\usepackage{lettrine}

\begin{document}
	
\begin{frontmatter}

\title{A Social Spider Algorithm for Solving the Non-convex Economic Load Dispatch Problem}

\author[hku]{James J.Q. Yu\corref{cor}}
\ead{jqyu@eee.hku.hk}
\author[hku]{Victor O.K. Li}
\ead{vli@eee.hku.hk}
\cortext[cor]{Corresponding author}
\address[hku]{Department of Electrical and Electronic Engineering, The University of Hong Kong, Pokfulam, Hong Kong}

\begin{abstract}
	Economic Load Dispatch (ELD) is one of the essential components in power system control and operation. Although conventional ELD formulation can be solved using mathematical programming techniques, modern power system introduces new models of the power units which are non-convex, non-differentiable, and sometimes non-continuous. In order to solve such non-convex ELD problems, in this paper we propose a new approach based on the Social Spider Algorithm (SSA). The classical SSA is modified and enhanced to adapt to the unique characteristics of ELD problems, e.g., valve-point effects, multi-fuel operations, prohibited operating zones, and line losses. To demonstrate the superiority of our proposed approach, five widely-adopted test systems are employed and the simulation results are compared with the state-of-the-art algorithms. In addition, the parameter sensitivity is illustrated by a series of simulations. The simulation results show that SSA can solve ELD problems effectively and efficiently.
\end{abstract}

\begin{keyword}
	Economic load dispatch, Social spider algorithm, non-convex optimization, valve-point effect.
\end{keyword}

\end{frontmatter}

\section{Introduction}
\lettrine[lines=2]{E}{conomic} Load Dispatch (ELD) is a fundamental problem in power system control and operation. The goal for ELD is to find a best feasible power generation schedule with a minimal fuel cost, while satisfying the generation constraints of the power units \cite{WoodWollenberg1984PowerGenerationOperation}. In the canonical formulation of ELD, the fuel costs of power units are represented by quadratic functions, which are convex and can be easily solved using mathematical programming methods. Many classical methods have been employed to solve ELD in the past decades, e.g., the gradient method \cite{DoduMartinMerlinPouget1972optimalformulationand}, the lambda iteration method \cite{ChenWang1993Branchandbound}, and quadratic programming \cite{FanZhang1998Realtimeeconomic}. These methods have also been employed to solve other optimization problems in power system like the Unit Commitment problem \cite{YuLiLam2013OptimalV2GScheduling} and the Optimal Power Flow problem \cite{JanChen1995ApplicationfastNewton}.

Although the convex, differentiable, and monotonically increasing canonical formulation of ELD is simple to solve, it is unrealistic because valve-point effects (VPE), multi-fuel options (MFO), and prohibited operating zones (POZ) are not considered. However, all these factors shall be accounted for in the real-world industrial production process. Incorporating these factors, the modern ELD is represented by a non-convex, non-continuous, and non-differentiable optimization problem with many equality and inequality constraints, making it very challenging to find the global optimum solution. For the sake of simplicity, ELD is used to refer to the modern formulation of the problem hereafter.

Despite the complexity of the problem, a number of techniques have been devised to solve ELD in the past decade, e.g., Tabu search \cite{LinChengTsay2002improvedtabusearch}, Taguchi method \cite{LiuCai2005Taguchimethodsolving}, and variants of particle swarm optimization \cite{Gaing2003Particleswarmoptimization,ParkJeongShinLee2010ImprovedParticleSwarm}. Evolutionary algorithms (EAs) also play an important role in solving ELD problems. Currently most state-of-the-art solvers for ELD are EAs and their variants according to the analysis in \cite{ZhanWuGuoZhou2015EconomicDispatchWith}.

Social Spider Algorithm (SSA) is a recently proposed evolutionary algorithm to solve global numerical optimization problems \cite{YuLi2015SocialSpiderAlgorithm}. By mimicking the foraging behavior of the social spiders, SSA explores and exploits the solution space in an iterative manner. In the formulation of SSA, searching information is propagated among the individuals, i.e., spiders, through the means of vibrations, which are lossy. In addition to this lossy information feature, SSA also incorporates a new social animal foraging model, namely, the information sharing model \cite{ClarkMangel1984ForagingandFlocking}. In this model, individuals in a population perform searching and joining behaviors simultaneously, which could potentially result in improved searching efficiency \cite{YuLi2015SocialSpiderAlgorithm,Uetz1992Foragingstrategiesspiders}. The reasons leading to the outstanding performance of SSA have been investigated in \cite{YuLi2015SocialSpiderAlgorithm}, and the improvements are mainly credited to the unique design of the information loss scheme and the searching pattern. Besides its superiority in solving optimization benchmark problems \cite{YuLi2015SocialSpiderAlgorithm}, SSA has also demonstrated its potential to be applied to address real world complex optimization problems \cite{YuLi2014BaseStationSwitching}. This makes it a good candidate to generate outstanding power schedules for ELD.

In this paper, we propose a variant of SSA to solve ELD problem, accounting for VPE, MFO, POZ, and power line loss. The advantage of our proposed algorithm is that it can generate more cost-efficient power schedules when compared with other algorithms. The rest of the paper is organized as follows. We first introduce the related work in Section \ref{sec:related}. Section \ref{sec:eld} presents the formulation of ELD with VPE, MFO, POZ, and power line loss. Our proposed algorithm is elaborated in Section \ref{sec:ssa}, and simulation problems, results, and comparisons are shown in Section \ref{sec:simr}. Finally we conclude this paper in Section \ref{sec:conclusion}.

\section{Related Work}\label{sec:related}

Over the past decades, many methods have been developed to solve ELD. Lin and Viviani proposed a hierarchical numerical method to solve the economic dispatch problem with piecewise quadratic cost functions \cite{LinViviani1984HierarchicalEconomicDispatch}. In this work the authors considered multiple intersecting cost functions for each generator, which is an analogy of MFO. A similar formulation of the problem is addressed by Park \textit{et al.} \cite{ParkKimEomLee1993Economicloaddispatch} with hopfield neural networks. This work is among the first attempts of adopting computational intelligence methodologies in solving ELD. Lee \textit{et al.} later proposed an adaptive hopfield neural network to solve the same problem \cite{LeeSode-YomePark1998AdaptiveHopfieldneural}. Their algorithm introduced a slope adjustment and bias adjustment method to speed up the convergence of the hopfield neural network system with adaptive learning rates. Lee and Breipohl proposed a decomposition technique to solve ELD with POZ \cite{LeeBreipohl1993Reserveconstrainedeconomic}. Their algorithm decomposes the nonconvex decision space into subsets which can be solved via the conventional Lagrangian relaxation approach. Binetti proposed a distributed algorithm based on the auction techniques and consensus protocols to solve ELD \cite{BinettiDavoudiNasoTurchianoLewis2014DistributedAuctionBased}. In their work, each power unit locally evaluates its possible fuel costs as bids. The bids are later employed in the auction mechanism to come up with a consensus. A very recent work by Zhan \textit{et al.} proposed a dimensional steepest decline method \cite{ZhanWuGuoZhou2015EconomicDispatchWith}. This method utilizes the local minimum analysis of the ELD problem to reduce the solution space to singular points.

Besides the above non-EA approaches, many EA methods have also been developed to solve various formulations of ELD. Orero and Irving proposed a simple Genetic Algorithm (GA) to solve ELD with POZ \cite{OreroIrving1996Economicdispatchgenerators}. Besides the standard GA, this work also devised a deterministic crowding GA model to solve the problem. Chiang developed an improved GA with the multiplier updating scheme for ELD with VPE and MFO \cite{Chiang2005Improvedgeneticalgorithm}. In this work, the proposed GA is incorporated with an improved evolutionary direction operator. In addition, the tailor-made migration operator efficiently searches the solution space. He \textit{et al.} proposed a hybrid GA approach to solve ELD with VPE \cite{HeWangMao2008hybridgeneticalgorithm}. The algorithm proposed is a hybrid GA with differential evolution (DE) and sequential quadratic programming (SQP). Sinha \textit{et al.} developed an Evolutionary Programming (EP) method to solve ELP with VPE \cite{SinhaChakrabartiChattopadhyay2003Evolutionaryprogrammingtechniques}. Pereira-Neto \textit{et al.} proposed an Evolutionary Strategy (ES) method to solve ELP with VPE and POZ \cite{Pereira-NetoUnsihuaySaavedra2005Efficientevolutionarystrategy}. DE has also been adapted to solve ELD \cite{WangChiouLiu2007Nonsmoothnonconvex,NomanaIb2008Differentialevolutioneconomic}.

Swarm Intelligence (SI), a branch of EA, has also attracted researchers' attention. Particle Swarm Optimization (PSO) has made a significant contribution in solving ELD problems. Selvakumar and Thanushkodi proposed a ``new PSO'' based on the classical PSO for ELD with VPE, MFO, and POZ \cite{SelvakumarThanushkodi2007NewParticleSwarm}. They manipulated the cognitive searching behavior in PSO to facilitate the solution space exploration. They also proposed an anti-predatory PSO in \cite{SelvakumaraThanushkodi2008Antipredatoryparticle}. In this algorithm, a new anti-predator scheme is modeled and introduced in the classical PSO. Chaturvedi \textit{et al.} proposed a hierarchical PSO for ELD with VPE and POZ \cite{ChaturvediGwaliorPanditSrivastava2008SelfOrganizingHierarchical}. In this work, a time-varying acceleration coefficient is introduced to act as the inertia factor of PSO. Meng \textit{et al.} proposed a Quantum PSO for ELD with VPE \cite{MengWangDongWong2010QuantumInspiredParticle}. Their algorithm demonstrated strong searching ability and fast convergence speed, which are contributed by the introduction of quantum computing theory, self-adaptive probability selection, and chaotic sequence mutation. Safari and Shayeghi developed an Iteration PSO for ELD with VPE and POZ \cite{SafariaShayeghi2011Iterationparticleswarm}. Besides the conventional global best (\textit{gBest}) and personal best (\textit{pBest}) positions considered in canonical PSO, the proposed algorithm also considers an iteration best (\textit{iBest}) position in the searching process. Nature-inspired EAs also yield satisfactory results in solving ELD variants. Some outstanding ones are Bee Colony Optimization Algorithm \cite{KumaraSaduaKumaraPanda2012novelmultiobjective}, Biogeography-Based Optimization \cite{BhattacharyaChattopadhyay2010BiogeographyBasedOptimization}, Ant Swarm Optimization \cite{CaiLiLiPengYang2012fuzzyadaptivechaotic}, Harmony Search Algorithm \cite{SantosMariani2009improvedharmonysearch}, and Chemical Reaction Optimization \cite{RoyBhuiPaul2014Solutioneconomicload}.

\section{Economic Load Dispatch Problem}\label{sec:eld}

The objective of the ELD problem is to find an optimal power generation schedule with minimal fuel cost while satisfying different power system operating constraints, including power unit and load balancing constraints. In this paper we adopt the formulation described in \cite{ZhanWuGuoZhou2015EconomicDispatchWith} and \cite{RoyBhuiPaul2014Solutioneconomicload}. The problem is formulated on one-hour time spans.

\subsection{Objective Function}

The objective function of ELD is defined as follows:
\begin{equation}\label{eqn:eld}
	\min_P \sum^n_{i=1}F_i^c(P_i),
\end{equation}
where $n$ is the total number of power units, $F_i^c(P_i)$ is the fuel cost function for the $i$th power unit, and $P_i$ is the power generation for the $i$th power unit according to the power generation schedule.

\subsubsection{Valve-point Effect}

Conventionally the fuel cost of power units are formulated by a quadratic function with the following form:
\begin{equation}
	F_i^c = a_i + b_iP_i + c_iP_i^2,
\end{equation}
where $a$, $b$, and $c$ are constant coefficients determined by the physical characteristics of the power units. However, the fuel cost function exhibits a larger variation in practice due to VPE, which generates ripple like effect during the valve-opening process of multi-valve units. A more precise formulation with both a quadratic component and a rectified sinusoidal component is adopted. In (\ref{eqn:eld}), the fuel cost is defined by
\begin{equation}\label{eqn:vpe}
	F_i^c = a_i + b_iP_i + c_iP_i^2 + |e_i\sin(f_i(P^{\textit{min}}_i)-P_i)|,
\end{equation}
where $e$ and $f$ are new coefficients describing VPE, and $P^{\textit{min}}_i$ is the minimum power generation for the $i$th power unit in the system.

\subsubsection{Multi-fuel Options}

Modern power units can be operated with multiple fuels \cite{ZhanWuGuoZhou2015EconomicDispatchWith}, and each fuel has a different fuel cost function. The unit will always utilize the fuel with a minimum fuel cost given a specified power generation requirement. Thus the fuel cost defined in (\ref{eqn:vpe}) is further modified to reflect the effects of multiple fuel options. A piecewise quadratic function is adopted to calculate the fuel cost of such power units, defined as follows:
\begin{align}\label{eqn:mfo}
	F_i^c = 
	\min(&a_{i,1} + b_{i,1}P_i + c_{i,1}P_i^2 + |e_{i,1}\sin(f_{i,1}(P^{\textit{min}}_i)-P_i)|, \notag\\
				 &a_{i,2} + b_{i,2}P_i + c_{i,2}P_i^2 + |e_{i,2}\sin(f_{i,2}(P^{\textit{min}}_i)-P_i)|, \notag\\
				 &\cdots, \notag\\
				 &a_{i,h} + b_{i,h}P_i + c_{i,h}P_i^2 + |e_{i,h}\sin(f_{i,h}(P^{\textit{min}}_i)-P_i)|),
\end{align}
where $a_{i,k}$, $b_{i,k}$, $c_{i,k}$, $e_{i,k}$, and $f_{i,k}$ are the fuel cost coefficients of the $k$th fuel option of the $i$th power unit, and $h$ is the total number of fuel options. Note that our formulation of MFO is different from the previous ones \cite{ZhanWuGuoZhou2015EconomicDispatchWith,RoyBhuiPaul2014Solutioneconomicload}, in which predefined power levels of switching among fuel options are listed as follows:
\begin{align}\label{eqn:mfo_old}
F_i^c =
\begin{cases}
a_{i,1} + b_{i,1}P_i + c_{i,1}P_i^2 + |e_{i,1}\sin(f_{i,1}(P^{\textit{min}}_i)-P_i)|& \text{if }P_i\in [P^{\textit{min}}_i, P_1) \\
a_{i,2} + b_{i,2}P_i + c_{i,2}P_i^2 + |e_{i,2}\sin(f_{i,2}(P^{\textit{min}}_i)-P_i)|& \text{if }P_i\in [P_1, P_2) \\
\cdots\\
a_{i,h} + b_{i,h}P_i + c_{i,h}P_i^2 + |e_{i,h}\sin(f_{i,h}(P^{\textit{min}}_i)-P_i)|& \text{if }P_i\in [P_{h-1}, P^{\textit{max}}_i]
\end{cases},
\end{align}
where $P^{\textit{max}}_i$ is the maximum power generation for the $i$th power unit, and $P_1,P_2,\cdots,P_{h-1}$ are the predefined power levels of switching fuel options. As all the formulations considering MFO makes the assumption that power units can choose fuel options freely \textit{a-priori}, our formulation is a more practical one. Meanwhile, as the predefined power levels in the simulation cases presented in Section \ref{sec:simr} were previously manipulated to make (\ref{eqn:mfo}) and (\ref{eqn:mfo_old}) equivalent, it is still fair to compare the performance of our proposed algorithm with existing ones. For example, $P_1$ value given in the previous test cases makes
\begin{align}
F_i^c = & a_{i,1} + b_{i,1}P_1 + c_{i,1}P_1^2 + |e_{i,1}\sin(f_{i,1}(P^{\textit{min}}_i)-P_1)| \notag\\
= & a_{i,2} + b_{i,2}P_1 + c_{i,2}P_1^2 + |e_{i,2}\sin(f_{i,2}(P^{\textit{min}}_i)-P_1)|.
\end{align}
Thus it is equivalent to
\begin{align}
F_i^c = 
\min(&a_{i,1} + b_{i,1}P_1 + c_{i,1}P_1^2 + |e_{i,1}\sin(f_{i,1}(P^{\textit{min}}_i)-P_1)|, \notag\\
&a_{i,2} + b_{i,2}P_1 + c_{i,2}P_1^2 + |e_{i,2}\sin(f_{i,2}(P^{\textit{min}}_i)-P_1)|).
\end{align}

\subsection{Constraints}
\subsubsection{Active Power Balance}

In this formulation we take the transmission line loss into consideration. Thus the active power balance is defined as an equilibrium between generated power and load demand plus line loss:
\begin{equation}\label{eqn:const_start}
	\sum^n_{i=1}P_i = P^{\textit{demand}} + P^{\textit{loss}},
\end{equation}
where $P^{\textit{demand}}$ and $P^{\textit{loss}}$ are the load demand and line loss, respectively. The line loss is calculated by \cite{WoodWollenberg1984PowerGenerationOperation}:
\begin{equation}
	P^{\textit{loss}} = \sum^n_{i=1}\sum^n_{j=1}P_iB_{ij}P_j+\sum^n_{i=1}B_{i0}P_i+B_{00},
\end{equation}
where $B_{ij}$ are the line loss coefficients.

\subsubsection{Power Generation Constraint}

The amount of power that each power unit can generate is limited by three factors: power limits, ramp rate limits, and POZ, each of which is represented by a set of inequalities. First, the power generation shall be within each power unit's minimum and maximum limits:
\begin{equation}\label{eqn:const_gen}
	P^{\textit{min}}_i\leq P_i\leq P^{\textit{max}}_i,
\end{equation}
where $P^{\textit{min}}_i$ and $P^{\textit{max}}_i$ are the minimum and maximum power output of the $i$th power unit, respectively. Second, ramp rates are employed to prevent severe power output changes, which are actually restricted by the physical properties of the power units. The operating ranges of all units are limited by their corresponding ramp rates:
\begin{equation}\label{eqn:ramp_rate}
	P^{\textit{prev}}_i-P^{\textit{DR}}\leq P_i\leq P^{\textit{prev}}_i+P^{\textit{UR}},
\end{equation}
where $P^{\textit{prev}}_i$ is the previous power output of the $i$th power unit, $P^{\textit{DR}}$ and $P^{\textit{UR}}$ are the ramp down and ramp up limits, respectively. Sometimes the entire operating range may not be completely feasible to the power unit due to physical operation limitations. Such power units have one or multiple power output ranges that are forbidden to the units. Therefore, additional constraints are introduced for the power units with POZ:
\begin{equation}\label{eqn:const_stop}
	P_i\in[P^{\textit{min}}_i, P^{\textit{l,1}}_i]\cup[P^{\textit{u,1}}_i, P^{\textit{l,2}}_i]\cup\cdots\cup[P^{\textit{u,z}}_i, P^{\textit{max}}_i],
\end{equation}
where $P^{\textit{l,r}}_i$ and $P^{\textit{u,r}}_i$ are the $r$th POZ of the $i$th power unit, and $z$ is the total number of POZs.

\section{Proposed Approach Based on Social Spider Algorithm}\label{sec:ssa}

SSA was recently proposed by Yu and Li \cite{YuLi2015SocialSpiderAlgorithm} to solve global numerical optimization problems. It is a general-purpose swarm intelligence algorithm utilizing the foraging behavior of the social spiders to perform optimization tasks. SSA was initially designed to solve continuous unconstrained problems, and we made several essential modifications to adapt the algorithm to solve ELD efficiently.

\subsection{Spider}

Spiders are the basic operating agents of SSA. In SSA, the solution space of an optimization problem is formulated as a hyper-dimensional spider web $S$ on which the spiders can move freely. Each position on the web corresponds to a feasible solution to the optimization problem. The spider web also serves as a transmission media for the vibrations generated by the spiders.

Each spider, say the $i$th spider in the population, in SSA is characterized by two properties, namely, its position $P_i(t)\in S$ and fitness value $f(P_i(t))$, where $t$ is the current iteration index and $f(x)$ is the objective function. In addition, each spider holds several attributes which are utilized to guide its random walk process when searching the solution space. The searching pattern will be introduced later and these attributes are:
\begin{itemize}
	\item The target vibration $V_i^{\textit{tar}}$.
	\item The inactive degree $d_i^{\textit{in}}$.
	\item The movement in the previous iteration $|P_i(t)-P_i(t-1)|$.
	\item The dimension mask\footnote{A 0-1 binary vector of length equals to the dimension of the optimization problem.} $M_i$.
\end{itemize}

\subsection{Vibration}

The vibrations represent a key design of SSA. It distinguishes SSA from other swarm intelligence algorithms and incorporates the lossy information idea into meta-heuristic algorithm design.

According to observations, the spiders are extremely sensitive to the vibrations propagated over the spider web. They are able to distinguish different vibrations from all directions and tell their intensities \cite{Uetz1992Foragingstrategiesspiders}. In SSA design, a vibration will be generated whenever a spider performs an arbitrary movement. The vibration carries the optimization information of the corresponding spider, and is propagated to and received by others in the same population. In such a way, the population of spiders share their personal experience and form a communal knowledge of the solution space.

The vibrations in SSA are characterized by two properties, namely, its source location $L_i\in S$ and the intensity at its source $T_i\in[0,+\infty)$. In iteration $t$, whenever a spider moves to a new position $P_i(t)$, it generates a vibration at this position $L_i = P_i(t)$ with an intensity calculated based on the fitness value of the position\footnote{We only study minimization problems in this paper. In such cases, smaller objective values are translated into larger intensity values using (\ref{eqn:intensity}).}:
\begin{equation}\label{eqn:intensity}
	T_i = \log(\frac{1}{f(P_i(t)) - C}+1),
\end{equation}
where $C$ is a confidently small constant. The introduction of $C$ is to guarantee the feasibility of the $\log$ term in (\ref{eqn:intensity}). The values of $C$ will be elaborated later.

 A vibration, after being generated, will attenuate when propagated over the spider web. Thus upon receipt, the spiders can only get partial information of the vibration's source location and its attenuated intensity. The vibration attenuation process is defined as follows:
\begin{equation}\label{eqn:attenuation}
	T_i^{\textit{D}} = T_i\times\exp(-\frac{D}{\overline{\sigma}\times r_a}),
\end{equation}
where $T_i^{\textit{D}}$ is the attenuated intensity after being propagated over a distance $D$, $\overline{\sigma}$ is the mean of the standard deviation of the population's positions over all dimensions, and $r_a\in(0, +\infty)$ is the attenuation rate, which is a user-controlled parameter.

\subsection{Iteration}

The complete optimization process of SSA is divided into three phases: initialization, iteration, and final phase, where the most notable one is the iteration phase. This phase is constituted of several steps, namely, fitness evaluation, vibration processing, mask changing, random walk, and constraint handling.

Each iteration starts with the fitness evaluation step, where the fitness value of each spider in the population is evaluated and stored. It is worth mentioning that the fitness evaluation process is conducted once and only once per iteration.

After all fitness values are evaluated, each spider will generate a vibration at its current position using (\ref{eqn:intensity}). The vibrations are then propagated over the spider web using (\ref{eqn:attenuation}), and received by all other spiders. Upon receipt of all vibrations, each spider will select the one with the largest attenuated intensity, denoted by $V_i^{\textit{rcv}}$, and compare it with $V_i^{\textit{tar}}$. If $V_i^{\textit{rcv}}$ is larger, it is stored as the new $V_i^{\textit{tar}}$. In such cases the inactive degree $d_i^{\textit{in}}$ is reset to zero. Otherwise $V_i^{\textit{tar}}$ remains unchanged and $d_i^{\textit{in}}$ is incremented by one.

In SSA, the movements of spiders are guided by both $V_i^{\textit{tar}}$ and $M_i$. $M_i$ is manipulated after $V_i^{\textit{tar}}$ is determined. In this mask-changing step two user-controlled parameters, $p_c$ and $p_m$, are introduced and used to modify $M_i$. At the beginning of this step, each spider will decide whether its $M_i$ shall be changed, and the probability of changing is $1-{p_c}^{d_i^{\textit{in}}}$. If $M_i$ is determined to be modified, each bit of $M_i$ has a probability $p_m$ to be assigned with a one, and $1-p_m$ to be a zero. If all bits are set to zero, a random bit is changed to one in order to avoid getting stuck in local optima \cite{YuLi2015SocialSpiderAlgorithm}.

When all dimension masks are determined, each spider will perform a random walk, and then employ the constraint-handling scheme to repair the possible infeasible solutions generated in the previous step. These two steps are substantially modified to solve ELD, as elaborated below.

\subsection{Random Walk with Chaotic Sequence Based Memory Factor}

When an iteration of SSA proceeds to the random walk step, each spider shall hold updated $V_i^{\textit{tar}}$, $M_i$, and a collection of received vibrations. These pieces of information are utilized to construct a target position $P_i^\textit{tar}$ as follows:
\begin{equation}
	(P_i^\textit{tar})_j =
	\begin{cases}
	(V_i^{\textit{tar}})_j & \text{if }(M_i)_j = 0 \\
	(V_i^{\textit{rand}})_j & \text{if }(M_i)_j = 1 \\
	\end{cases},
\end{equation}
where $(P_i^\textit{tar})_j$ is the $j$th element of $P_i^\textit{tar}$, $(V_i^{\textit{tar}})_j$ is the $j$th element of the source location of $V_i^\textit{tar}$, $V_i^{\textit{rand}}$ is a random vibration received by the spider, and $(M_i)_j$ is the $j$th bit of $M_i$.

With the generated $P_i^\textit{tar}$, here we introduce a chaotic sequence based memory factor into the random walk process. Chaotic sequences has been employed in controlling the optimization process of many swarm intelligence algorithms \cite{ParkJeongShinLee2010ImprovedParticleSwarm,AraujoS.2008Particleswarmapproaches}. In our proposed algorithm, we employ a logistic map iterator to emulate the dynamic system with chaotic behavior \cite{CaponettoFortunaFazzinoXibilia2003Chaoticsequencesto}:
\begin{equation}
	\gamma_t = \mu\gamma_{t-1}(1-\gamma_{t-1}),
\end{equation}
where $\mu$ is a control parameter and is set to four in this paper. $\gamma_t$ is the chaotic parameter at iteration $t$, randomly generated in $(0.75,1)$ in this paper.

In addition, we introduce a new memory factor to control the impact of past behavior on the spider's random walk. In previous work \cite{YuLi2015SocialSpiderAlgorithm,YuLi2015ParameterSensitivityAnalysis}, the random walk formula is defined as follows:
\begin{equation}
	P_i(t+1) = P_i(t) + (P_i(t) - P_i(t-1))\times\delta+(P_i^\textit{tar} - P_i(t))\odot R,
\end{equation}
where $\delta$ is randomly generated in $(0,1)$, $R\sim U(0,1)$ is a vector of random numbers, and $\odot$ is the element-wise multiplication operation. In our formulation, $\delta$ is considered as the memory factor, and defined as follows:
\begin{equation}
	\delta(t) = \gamma_t(\omega^{\textit{max}}-\frac{\omega^{\textit{max}} - \omega^{\textit{min}}}{\textit{iter}^{\textit{max}}}\times t),
\end{equation}
where $\omega^{\textit{max}}$ and $\omega^{\textit{min}}$ are the maximum and minimum memory strengths, respectively. $\textit{iter}^{\textit{max}}$ is the maximum allowed iteration count, which is a stopping criteria. The design of $\omega$ terms is similar to the descending inertia weight approach used in PSO \cite{ParkJeongShinLee2010ImprovedParticleSwarm}.

\subsection{Power Schedule Repairing Scheme}

After the random walk step, the spiders in the population are assigned with new positions. However, as their positions are not checked against the constraints of ELD, namely, (\ref{eqn:const_start})--(\ref{eqn:const_stop}), an additional constraint-handling scheme shall be incorporated to repair the infeasible solutions.

We first consider the power generation constraints, i.e., (\ref{eqn:const_gen})--(\ref{eqn:const_stop}). In these constraints, several power levels are designed to limit the available power outputs: $P^{\textit{min}}_i$, $P^{\textit{max}}_i$, $P^{\textit{prev}}_i-P^{\textit{DR}}$,  $P^{\textit{prev}}_i+P^{\textit{UR}}$, $P^{\textit{l,r}}_i$, and $P^{\textit{u,r}}_i$. If an element $p_i$ in the checked power schedule (spider position) $P$ is infeasible for (\ref{eqn:const_gen})--(\ref{eqn:const_stop}), $p_i$ is set to the power level which is closest to the original $p_i$. Thus a boundary absorbing technique \cite{ChuGaoSorooshian2011Handlingboundaryconstraints} is employed to address (\ref{eqn:const_gen})--(\ref{eqn:const_stop}).

After all power outputs satisfy the boundary constraints, the active power balance constraint (\ref{eqn:const_start}) is checked. The deficit energy is calculated as follows:
\begin{equation}
	P^{\textit{dfc}} = P^{\textit{demand}} + P^{\textit{loss}} - \sum^n_{i=1}P_i.
\end{equation}
Then a repairing operation is repeated until $P^{\textit{dfc}} = 0$. The scheme first randomly selects the $g$th power unit and then calculates its remaining capacity:
\begin{equation}
	P^{\textit{cap}}_g = 
	\begin{cases}
	{P^{\textit{max}}_g}^* - p_g & P^{\textit{dfc}} > 0 \\
	{P^{\textit{min}}_g}^* - p_g & P^{\textit{dfc}} < 0 \\
	\end{cases},
\end{equation}
where ${P^{\textit{max}}_g}^*$ and ${P^{\textit{min}}_g}^*$ are the maximum and minimum power outputs in the current allowed operating zone. For the test instances without POZ, ${P^{\textit{max}}_g}^* = P^{\textit{max}}_g$ and ${P^{\textit{min}}_g}^* = P^{\textit{min}}_g$. Otherwise ${P^{\textit{max}}_g}^* = \min\{P^{\textit{max}}_g, P^{\textit{l,q}}_g\}$, where $P^{\textit{l,q}}_g$ is the closest lower limit of all POZs.
${P^{\textit{min}}_g}^* = \max\{P^{\textit{min}}_g, P^{\textit{u,q}}_g\}$, where $P^{\textit{u,q}}_g$ is the closest upper limit of all POZs. After $P^{\textit{cap}}_g$ is calculated, $p_g$ and $P^{\textit{dfc}}$ are manipulated according to the following:
\begin{align}
P^\textit{ch} =& 
\begin{cases}
\min\{P^{\textit{cap}}_g\times r, P^{\textit{dfc}}\} & P^{\textit{dfc}} > 0 \\
\max\{P^{\textit{cap}}_g\times r, P^{\textit{dfc}}\} & P^{\textit{dfc}} < 0 \\
\end{cases},\\
p_g \leftarrow& p_g + P^\textit{ch}, \\
P^{\textit{dfc}} \leftarrow& P^{\textit{dfc}} - P^\textit{ch}.
\end{align}

This repairing scheme will work for test instances without POZ. However, for some rare cases with POZ, it is possible that $\sum^n_{g=1}P^{\textit{cap}}_g$ is not sufficient to cover $P^{\textit{dfc}}$ due to the limitation of POZ. In such a case, the power output of a random unit is set to the closest upper limit of all POZs, i.e., $p_g=\min\{P^{\textit{u,q}}_g|P^{\textit{u,q}}_g>p_g\}$, and the repairing scheme is conducted again until constraint (\ref{eqn:const_start}) is satisfied. A pseudocode of the modified SSA with the proposed ELD-specific schemes is presented in Algorithm \ref{alg:ssa}.

\begin{algorithm}
	\caption{\sc{Modified Social Spider Algorithm for Economic Load Dispatch Problem}}
	\begin{algorithmic}[1]
		\State Assign values to the parameters of SSA.
		\State Create the population of spiders $pop$ and assign memory for them.
		\State Initialize $V_i^\textit{tar}$ for each spider.
		\While {stopping criteria not met}
		\For {\textbf{each} spider in $pop$}
		\State Evaluate the fitness value.
		\State Generate a vibration at the spider's position.
		\EndFor
		\For {\textbf{each} spider in $pop$}
		\State Calculate the intensity of the generated vibrations.
		\State Select the strongest vibration $V_i^\textit{rcv}$.
		\If {The intensity of $V_i^\textit{rcv}$ is larger than $V_i^\textit{tar}$}
		\State Store $V_i^\textit{rcv}$ as $V_i^\textit{tar}$.
		\EndIf
		\State Update $d_i^{in}$.
		\State Generate a random number $r$ from [0,1).
		\If {$r > {p_c}^{d_i^{in}}$}
		\State Update the dimension mask $M_i$.
		\EndIf
		\State Update the logistic map iterator $\gamma_t$.
		\State Update the memory factor $\delta(t)$ with $\gamma_t$.
		\State Perform a random walk with $\delta(t)$.
		\State Address violated constraints in the power schedule using proposed repairing scheme.
		\EndFor
		\EndWhile
		\State Output the best solution found.
	\end{algorithmic}
	\label{alg:ssa}
\end{algorithm}

\section{Simulation Results and Comparisons}\label{sec:simr}

In order to benchmark the performance of our proposed SSA-based approach in solving variants of ELD, we conduct a series of simulations on test systems with different combinations of VPE, MFO, POZ, and line loss. Five different test cases are considered, where the time span for each case is one hour \cite{SinhaChakrabartiChattopadhyay2003Evolutionaryprogrammingtechniques}. In addition, the ramp rate constraint (\ref{eqn:ramp_rate}) is provided in the last two cases. For all simulations, the population size equals the number of power units, $r_a$ is set to 10, $p_c$ and $p_m$ are set to 0.9 and 0.1, respectively. Constant $C$ in (\ref{eqn:intensity}) is set to the value of the minimal fuel cost $\sum^n_{i=1}F_i^c(P^{\textit{min}}_i)$. The stopping criteria for all simulations is 100000 function evaluations, and each test system is repeated for 25 runs.

The algorithm is implemented in C++. All simulations of SSA are conducted on a personal computer with an Intel Core i7 CPU at 3.40GHz. For other algorithms, the best power schedules are obtained from the corresponding publications and the fuel costs are evaluated using these obtained solutions.

\subsection{ELD with VPE}\label{sys:1}

For this variant of ELD, two test systems are adopted for evaluation, namely, the 13-unit system \cite{SinhaChakrabartiChattopadhyay2003Evolutionaryprogrammingtechniques} and the 40-unit system \cite{SinhaChakrabartiChattopadhyay2003Evolutionaryprogrammingtechniques}. Load demand for these systems are 1800MW and 10 500MW. The system coefficients are presented in Tables \ref{tbl:13unit_config} and \ref{tbl:40unit_config}.

The performance of SSA is compared with the state-of-the-art algorithms in solving these two test systems, namely, Quantum PSO (QPSO) \cite{MengWangDongWong2010QuantumInspiredParticle}, hybrid GA (HGA) \cite{HeWangMao2008hybridgeneticalgorithm}, iteration PSO with time varying acceleration coefficients (IPSO-TVAC) \cite{SafariaShayeghi2011Iterationparticleswarm}, self-tuning hybrid DE (SHDE) \cite{WangChiouLiu2007Nonsmoothnonconvex}, fuzzy adaptive PSO with variable DE (FAPSO-VDE) \cite{NiknamMojarradMeym2011novelhybridparticle}, hybrid CRO with DE (HCRO-DE) \cite{RoyBhuiPaul2014Solutioneconomicload}, Dimensional Steepest Decline Method (DSD) \cite{ZhanWuGuoZhou2015EconomicDispatchWith}, and PSO with chaotic sequences and crossover operation (CCPSO) \cite{ParkJeongShinLee2010ImprovedParticleSwarm}. It should be noted that the total cost obtained can be different from the publication due to calculation precision issues \cite{ZhanWuGuoZhou2015EconomicDispatchWith}.

\begin{table*}
	\caption{Simulation Results for 13-unit Test System with VPE}
	\label{tbl:13unit}
	\scriptsize
	\begin{center}
		\begin{tabular}{crrrrrrrr}
			\hline
			Unit & SSA & QPSO & HGA & IPSO-TVAC & SHDE & FAPSO-VDE & HCRO-DE & DSD \\ 
			\hline
			1 &  628.31788 & 538.56 & 628.3185 & 628.3185 & 628.3172 & 628.3185 & 628.3185 & 628.31853 \\
			2 &  149.57315 & 224.70 & 222.7491 & 149.5996 & 149.5986 & 222.7490 & 149.5930 & 149.59965 \\
			3 &  224.38835 & 150.09 & 149.5996 & 222.7489 & 222.7987 & 149.5990 & 222.7559 & 222.74907 \\
			4 &  109.86655 & 109.87 & 109.8665 & 109.8666 & 109.8673 & 109.8665 & 109.8665 & 109.86655 \\
			5 &  109.86652 & 109.87 & 109.8665 & 109.8666 & 109.8418 & 109.8665 & 109.8665 & 109.86655 \\
			6 &  109.86592 & 109.87 & 109.8665 & 109.8666 & 109.8641 & 109.8665 & 109.8665 & 109.86655 \\
			7 &  109.86439 & 109.87 & 109.8665 & 109.8666 & 109.8547 & 109.8665 & 109.8665 & 109.86655 \\
			8 &  109.86644 & 159.75 & 109.8665 & 109.8666 & 109.8576 & 109.8665 & 109.8665 & 109.86655 \\
			9 &   60.00000 & 109.87 &  60.0000 &  60.0000 &  60.0000 &  60.0000 &  60.0000 &  60.00000 \\
			10 &  40.00000 &  77.41 &  40.0000 &  40.0000 &  40.0000 &  40.0000 &  40.0000 &  40.00000 \\
			11 &  40.00000 &  40.00 &  40.0000 &  40.0000 &  40.0000 &  40.0000 &  40.0000 &  40.00000 \\
			12 &  55.00000 &  55.01 &  55.0000 &  55.0000 &  55.0000 &  55.0000 &  55.0001 &  55.00000 \\
			13 &  55.00000 &  55.01 &  55.0000 &  55.0000 &  55.0000 &  55.0000 &  55.0000 &  55.00000 \\
			Cost & \textbf{17963.766} & 18398.848 & 17963.829 & 17963.833 & 17963.891 & 17963.829 & 17963.831 & 17963.829 \\
			\hline
		\end{tabular}
	\end{center}
\end{table*}

\begin{table*}
	\caption{Simulation Results for 40-unit Test System with VPE}
	\label{tbl:40unit}
	\scriptsize
	\begin{center}
		\begin{tabular}{crrrrrrrr}
			\hline
			Unit & SSA & QPSO & HGA & IPSO-TVAC & FAPSO-VDE & HCRO-DE & DSD & CCPSO \\ 
			\hline
			1 & 110.80000 & 111.20 & 111.3793 & 110.80 & 110.8018 & 110.8015 & 110.79983 & 110.7998 \\
			2 & 110.80000 & 111.70 & 110.9278 & 110.80 & 110.8000 & 110.7998 & 110.79983 & 110.7999 \\
			3 & 97.50000 & 97.40 & 97.4104 & 97.40 & 97.3999 & 97.3999 & 97.39991 & 97.3999 \\
			4 & 179.69999 & 179.73 & 179.7331 & 179.73 & 179.7331 & 179.7331 & 179.73310 & 179.7331 \\
			5 & 87.79992 & 90.14 & 89.2188 & 87.80 & 87.7998 & 87.7999 & 87.79990 & 87.7999 \\
			6 & 140.00000 & 140.00 & 140.0000 & 140.00 & 140.0000 & 140.0000 & 140.00000 & 140.0000 \\
			7 & 259.59973 & 259.60 & 259.6198 & 259.60 & 259.5997 & 259.5997 & 259.59965 & 259.5997 \\
			8 & 284.59980 & 284.80 & 284.6570 & 284.60 & 284.5997 & 284.5997 & 284.59965 & 284.5997 \\
			9 & 284.59957 & 284.84 & 284.6588 & 284.60 & 284.5997 & 284.5997 & 284.59965 & 284.5997 \\
			10 & 130.00000 & 130.00 & 130.0000 & 130.00 & 130.0000 & 130.0000 & 130.00000 & 130.0000 \\
			11 & 94.00000 & 168.80 & 168.8214 & 94.00 & 94.0000 & 94.0000 & 94.00000 & 94.0000 \\
			12 & 94.00000 & 168.80 & 168.8496 & 94.00 & 94.0000 & 94.0000 & 94.00000 & 94.0000 \\
			13 & 214.75979 & 214.76 & 214.7524 & 214.76 & 214.7598 & 214.7598 & 214.75979 & 214.7598 \\
			14 & 394.27937 & 304.53 & 394.2848 & 394.28 & 394.2794 & 394.2794 & 394.27937 & 394.2794 \\
			15 & 394.27937 & 394.28 & 304.5361 & 394.28 & 394.2794 & 394.2794 & 394.27937 & 394.2794 \\
			16 & 394.27937 & 394.28 & 394.2987 & 394.28 & 394.2794 & 394.2794 & 394.27937 & 394.2794 \\
			17 & 489.27937 & 489.28 & 489.2877 & 489.28 & 489.2794 & 489.2794 & 489.27937 & 489.2794 \\
			18 & 489.27937 & 489.28 & 489.2869 & 489.28 & 489.2794 & 489.2794 & 489.27937 & 489.2794 \\
			19 & 511.27937 & 511.28 & 511.2752 & 511.28 & 511.2794 & 511.2794 & 511.27937 & 511.2794 \\
			20 & 511.27937 & 511.28 & 511.2857 & 511.28 & 511.2794 & 511.2794 & 511.27937 & 511.2794 \\
			21 & 523.27937 & 523.28 & 523.2961 & 523.28 & 523.2794 & 523.2794 & 523.27937 & 523.2794 \\
			22 & 523.27937 & 523.28 & 523.3202 & 523.28 & 523.2807 & 523.2794 & 523.27937 & 523.2794 \\
			23 & 523.27937 & 523.29 & 523.2916 & 523.28 & 523.0000 & 523.2794 & 523.27937 & 523.2794 \\
			24 & 523.27937 & 523.28 & 523.3014 & 523.28 & 523.0000 & 523.2794 & 523.27937 & 523.2794 \\
			25 & 523.27937 & 523.29 & 523.2675 & 523.28 & 523.0000 & 523.2794 & 523.27937 & 523.2794 \\
			26 & 523.27937 & 523.28 & 523.2787 & 523.28 & 523.0000 & 523.2790 & 523.27937 & 523.2794 \\
			27 & 10.00000 & 10.01 & 10.0000 & 10.00 & 10.0000 & 10.0000 & 10.00000 & 10.0000 \\
			28 & 10.00000 & 10.01 & 10.0000 & 10.00 & 10.0000 & 10.0000 & 10.00000 & 10.0000 \\
			29 & 10.00000 & 10.00 & 10.0000 & 10.00 & 10.0000 & 10.0000 & 10.00000 & 10.0000 \\
			30 & 87.80000 & 88.47 & 88.6376 & 87.80 & 87.7999 & 87.7999 & 87.79990 & 87.8000 \\
			31 & 190.00000 & 190.00 & 190.0000 & 190.00 & 190.0000 & 190.0000 & 190.00000 & 190.0000 \\
			32 & 190.00000 & 190.00 & 190.0000 & 190.00 & 190.0000 & 190.0000 & 190.00000 & 190.0000 \\
			33 & 190.00000 & 190.00 & 190.0000 & 190.00 & 190.0000 & 190.0000 & 190.00000 & 190.0000 \\
			34 & 164.68395 & 164.91 & 164.9795 & 164.80 & 164.8015 & 164.7998 & 164.79983 & 164.7998 \\
			35 & 194.44082 & 165.36 & 165.9970 & 194.40 & 194.3928 & 194.3956 & 194.39778 & 194.3976 \\
			36 & 200.00000 & 167.19 & 165.0464 & 200.00 & 200.0000 & 200.0000 & 200.00000 & 200.0000 \\
			37 & 110.00000 & 110.00 & 110.0000 & 110.00 & 110.0000 & 110.0000 & 110.00000 & 110.0000 \\
			38 & 110.00000 & 107.01 & 110.0000 & 110.00 & 110.0000 & 110.0000 & 110.00000 & 110.0000 \\
			39 & 110.00000 & 110.00 & 110.0000 & 110.00 & 110.0000 & 110.0000 & 110.00000 & 110.0000 \\
			40 & 511.28462 & 511.36 & 511.3005 & 511.28 & 511.2794 & 511.2794 & 511.27937 & 511.2794 \\
			Cost & 121412.55 & 121448.21 & 121418.27 & 121412.54 & 121412.56 & 121412.55 & \textbf{121412.53} & 121412.54 \\
			\hline
		\end{tabular}
	\end{center}
\end{table*}

Simulation results and best power schedules are listed in Tables \ref{tbl:13unit} and \ref{tbl:40unit}. From the tables we can see that all constraints are satisfied with the power schedule found, and the obtained solutions of the best performing algorithms (SSA, DSD, HGA, and FAPSO-VDE for 13-unit system; SSA, IPSO-TVAC, DSD, and CCSPO for 40-unit system) are quite similar. SSA outperforms others in 13-unit comparison and achieves a satisfactory performance in 40-unit system compared with other state-of-the-art algorithms. These results demonstrates the superiority of SSA in exploiting the local optimum spaces.

The average computation times for SSA are 0.474 and 0.569 seconds for the 13 and 40-unit systems, respectively. As a reference, the corresponding simulation times for FAPSO-VDE were 4.1 and 22 seconds with an Intel Pentium IV CPU at 3.0GHz \cite{NiknamMojarradMeym2011novelhybridparticle}, which can be roughly translated into approximately 2 and 12 seconds on our simulation platform, respectively. Typically ELD is usually considered with the Unit Commitment Problem, which makes power generation schedules on a per-hour basis. In such cases a computational time of several seconds is not as significant when compared with the optimization time frame. However, with the introduction of fast-changing renewable energy sources to the conventional grid, in the future, a smaller response time for such generation optimization problems will be required and it will be necessary to achieve a faster computational speed.

\subsection{ELD with VPE and MFO}

For ELD with VPE and MFO, we employ a 10-unit system \cite{Chiang2005Improvedgeneticalgorithm} for comparison. Load demand for this system is 2700MW. The system coefficients are presented in Table \ref{tbl:10unit_config}.

The performance of SSA is compared with the state-of-the-art algorithms in solving this test system, namely, QPSO, PSO with local random search (PSO-LRS) \cite{SelvakumarThanushkodi2007NewParticleSwarm}, new PSO with local random search (NPSO-LRS) \cite{SelvakumarThanushkodi2007NewParticleSwarm}, PSO with the constriction factor and inertia weight with a real-valued mutation (CBPSO-RVM) \cite{LuSriyanyongSongDillon2010Experimentalstudynew}, improved GA
with multiplier updating (IGA-MU) \cite{Chiang2005Improvedgeneticalgorithm}, HCRO-DE, and DSD. The simulation results and comparison are presented in Table \ref{tbl:10unit}. The average computation time is 0.510 second. 

\begin{table*}
	\caption{Simulation Results for 10-unit Test System with VPE and MFO}
	\label{tbl:10unit}
	\scriptsize
	\begin{center}
		\begin{tabular}{crrrrrrrr}
			\hline
			Unit & SSA & QPSO & PSO\_LRS & NPSO\_LRS & CBPSO-RVM & IGA\_MU & HCRO-DE & DSD \\ 
			\hline1 & 219.16264 & 224.7063 & 219.0155 & 223.3352 & 219.2073 & 219.1261 & 213.4589 & 218.59400 \\
			2 & 211.65928 & 212.3882 & 213.8901 & 212.1957 & 210.2203 & 211.1645 & 209.7300 & 211.71174 \\
			3 & 280.68427 & 283.4405 & 283.7616 & 276.2167 & 278.5456 & 280.6572 & 332.0143 & 280.65706 \\
			4 & 239.95493 & 239.9530 & 237.2687 & 239.4187 & 239.3704 & 238.4770 & 237.7581 & 239.63943 \\
			5 & 276.38750 & 283.8190 & 286.0163 & 274.6470 & 276.4120 & 276.4179 & 269.1476 & 279.93452 \\
			6 & 239.79532 & 241.0024 & 239.3987 & 239.7974 & 240.5797 & 240.4672 & 238.9677 & 239.63943 \\
			7 & 290.07417 & 287.8671 & 291.1767 & 285.5388 & 292.3267 & 287.7399 & 280.6141 & 287.72749 \\
			8 & 239.82117 & 240.6245 & 241.4398 & 240.6323 & 237.7557 & 240.7614 & 238.9677 & 239.63943 \\
			9 & 426.37501 & 407.9870 & 416.9721 & 429.2637 & 429.4008 & 429.3370 & 413.6294 & 426.58829 \\
			10 & 276.08571 & 278.2120 & 271.0623 & 278.9541 & 276.1815 & 275.8518 & 266.3841 & 275.86861 \\
			Cost & \textbf{623.6433} & 624.1505 & 624.0297 & 623.9258 & 624.3911 & 623.6526 & 628.9605 & 623.8265 \\
			\hline
		\end{tabular}
	\end{center}
\end{table*}

From the simulation results it is clear that SSA again outperforms all compared algorithms in solving this ELD with VPE and MFO problem. In this comparison, the best power schedules found by the compared algorithms have some differences in terms of the outputs of the power units. A preliminary guess of the reason to this condition is that the solution space for ELD problems considering MFO is more complex than that with only VPE. This may potentially result in algorithms getting stuck in the local optima.

\subsection{ELD with POZ and Line Loss}

For ELD with POZ and line loss characteristics, we employ two test systems for comparison, namely, a 6-unit system \cite{Gaing2003Particleswarmoptimization} and a 15-unit system \cite{Gaing2003Particleswarmoptimization}. Load demand for these systems are 1263MW and 2630MW, respectively. The system coefficients are presented in Tables \ref{tbl:6unit_config} and \ref{tbl:15unit_config}, and line loss coefficients are listed in \cite{Gaing2003Particleswarmoptimization}.

The performance of SSA is compared with the state-of-the-art algorithms in solving these two test systems, namely, QPSO, NPSO-LRS, IPSO-TVAC, bacterial foraging optimization (BFO) \cite{PanigrahiPandi2008Bacterialforagingoptimisation}, improved PSO (IPSO) \cite{Gaing2003Particleswarmoptimization}, elitist GA (EGA) \cite{Gaing2003Particleswarmoptimization}, HCRO-DE, simple PSO (SPSO) \cite{ChaturvediGwaliorPanditSrivastava2008SelfOrganizingHierarchical},	passive congregation-based PSO (PC-PSO) \cite{ChaturvediGwaliorPanditSrivastava2008SelfOrganizingHierarchical}, self-organizing hierarchical PSO (SOH-PSO) \cite{ChaturvediGwaliorPanditSrivastava2008SelfOrganizingHierarchical}, firefly algorithm \cite{YangHosseiniGandomi2012FireflyAlgorithmsolving}, and CCPSO
. The simulation results and comparison are presented in Tables \ref{tbl:6unit} and \ref{tbl:15unit}. The average computation times are 0.338 and 0.574 second, respectively.

\begin{table*}
	\caption{Simulation Results for 6-unit Test System with POZ and Line Loss}
	\label{tbl:6unit}
	\scriptsize
	\begin{center}
		\begin{tabular}{crrrrrrrr}
			\hline
			Unit & SSA & QPSO & NPSO\_LRS & IPSO-TVAC & BFO & IPSO & EGA & HCRO-DE \\ 
			\hline
			1 & 448.39165 & 447.5823 & 446.9600 & 447.5840 & 449.4600 & 447.4970 & 474.8066 & 447.4021 \\
			2 & 169.30115 & 172.8387 & 173.3944 & 173.2010 & 172.8800 & 173.3221 & 178.6363 & 173.2407 \\
			3 & 256.19797 & 261.3300 & 262.3436 & 263.3310 & 263.4100 & 263.4745 & 262.2089 & 263.3812 \\
			4 & 139.74938 & 138.6812 & 139.5120 & 138.8520 & 143.4900 & 139.0594 & 134.2826 & 138.9774 \\
			5 & 170.27317 & 169.6781 & 164.7089 & 165.3280 & 164.9100 & 165.4761 & 151.9039 & 165.3897 \\
			6 & 89.72839 & 74.8963 & 89.0162 & 87.1500 & 81.2520 & 87.1280 & 74.1812 & 87.0538 \\
			Loss & \textbf{10.6421} & 13.0066 & 12.9351 & 12.4460 & 12.4020 & 12.9584 & 13.0217 & 12.4449 \\
			Cost & \textbf{15419.803} & 15450.140 & 15450.000 & 15443.063 & 15443.8497 & 15449.882 & 15459.239 & 15443.075 \\
			\hline
		\end{tabular}
	\end{center}
\end{table*}

\begin{table*}
	\caption{Simulation Results for 15-unit Test System with POZ and Line Loss}
	\label{tbl:15unit}
	\scriptsize
	\begin{center}
		\begin{tabular}{crrrrrrrr}
			\hline
			Unit & SSA & IPSO & EGA & SPSO & PC-PSO & SOH-PSO & FA & CCPSO\\ 
			\hline1 & 455.0000 & 439.1162 & 415.3108 & 455.00 & 455.00 & 455.00 & 455.0000 & 455.0000\\
			2 & 380.0000 & 407.9727 & 359.7206 & 380.00 & 380.00 & 380.00 & 380.0000 & 380.0000 \\
			3 & 130.0000 & 119.6324 & 104.4250 & 130.00 & 130.00 & 130.00 & 130.0000 & 130.0000 \\
			4 & 130.0000 & 129.9925 & 74.9853 & 129.28 & 127.15 & 130.00 & 130.0000 & 130.0000 \\
			5 & 169.9721 & 151.0681 & 380.2844 & 164.77 & 169.91 & 170.00 & 170.0000 & 170.0000 \\
			6 & 460.0000 & 459.9978 & 426.7902 & 460.00 & 460.00 & 459.96 & 460.0000 & 460.0000 \\
			7 & 430.0000 & 425.5601 & 341.3164 & 424.52 & 430.00 & 430.00 & 430.0000 & 430.0000 \\
			8 & 125.6909 & 98.5699 & 124.7867 & 60.00 & 108.38 & 117.53 & 71.7450 & 71.7526 \\
			9 & 32.5629 & 113.4936 & 133.1445 & 25.00 & 77.41 & 77.90 & 58.9164 & 58.9090 \\
			10 & 128.1047 & 101.1142 & 89.2567 & 160.00 & 97.76 & 119.54 & 160.0000 & 160.0000 \\
			11 & 80.0000 & 33.9116 & 60.0572 & 80.00 & 67.61 & 54.50 & 80.0000 & 80.0000 \\
			12 & 80.0000 & 79.9583 & 49.9998 & 72.62 & 73.26 & 80.00 & 80.0000 & 80.0000 \\
			13 & 25.0000 & 25.0042 & 38.7713 & 25.00 & 25.57 & 25.00 & 25.0000 & 25.0000 \\
			14 & 15.0000 & 41.4140 & 41.9425 & 44.38 & 19.57 & 17.86 & 15.0000 & 15.0000 \\
			15 & 15.0000 & 35.6140 & 22.6445 & 49.42 & 38.93 & 15.00 & 15.0000 & 15.0000 \\
			Loss & \textbf{26.3306} & 32.4196 & 33.4359 & 29.9930 & 30.5500 & 32.2900 & 30.6614 & 30.6616 \\
			Cost & \textbf{32662.51} & 32857.54 & 33063.54 & 32798.69 & 32775.36 & 32751.39 & 32704.45 & 32704.45\\
			\hline
		\end{tabular}
	\end{center}
\end{table*}

From the simulation results we can see SSA generates the best performing power schedule among all the compared algorithms. Further investigation of the generated schedules shows that although all algorithms can successfully locate the same best performing operating zones, other algorithms are not able to further exploit the optimum sub-space.

It is worth noting that there are several published results on the 15-unit system with better fuel cost performance, i.e., smaller than \$32 662.51 obtained by SSA. However, after a careful investigation it can be observed that these results violate the ramp rate constraints. For example, the power output of the second unit in the best recorded result of \cite{Gaing2003Particleswarmoptimization} is 407.9727, which exceeds the maximum allowed power output limited by the ramp rate, which is 380. This situation may be caused by the different test instance configurations, and these infeasible solutions are not included in the comparison (except for \cite{Gaing2003Particleswarmoptimization}, in which the test case was proposed).

\subsection{Parameter Selection}

Parameter selection is critical to the optimization performance of SSA \cite{YuLi2015ParameterSensitivityAnalysis}. Although there is already related work on benchmarking the parameter sensitivity of SSA, it is still interesting to investigate and search for the optimal combination of parameters to solve ELD-like optimization problems. In order to test the impact of changing parameters on the fuel cost performance, we employ the 13-unit test system introduced in Section \ref{sys:1} and perform a parameter sweep test on the four parameters of SSA, namely, population size ($|pop|$), $r_a$, $p_c$, and $p_m$.

The simulation is conducted as follows. We first start from the previous recommended parameter setting given in \cite{YuLi2015ParameterSensitivityAnalysis}, i.e., $|pop|$/$r_a$/$p_c$/$p_m$ is 30/1.0/0.7/0.1. Then one parameter is tested against a wide range of possible values to figure out which one performs the best. This process is repeated until all four parameters are adjusted. Note that although this testing method neglects the correlations among multiple parameters, it can still generate a sub-optimal parameter combination while alleviating the effort in the parameter-tuning process to the maximum extent.

\begin{table*}
	\caption{Parameter Analysis Simulation Results}
	\label{tbl:param}
	\scriptsize
	\begin{center}
		\begin{tabular}{cccc|cccc|cccc}
			\hline
			$r_a$ & Best & Mean & S.D. & $p_c$ & Best & Mean & S.D. & $p_m$ & Best & Mean & S.D. \\
			\hline
			0.1 & 17964.267 & 17964.669 & 0.3258 & 0.01 & 17964.136 & 17964.359 & 0.1312 & 0.01 & 17963.886 & 17963.946 & 0.0254 \\
			0.2 & 17964.187 & 17964.374 & 0.1059 & 0.1 & 17964.168 & 17964.388 & 0.1530 & 0.1 & \textbf{17963.864} & \textbf{17963.893} & \textbf{0.0142} \\
			0.5 & 17964.077 & 17964.229 & 0.1026 & 0.2 & 17964.119 & 17964.209 & 0.0580 & 0.2 & 17963.944 & 17964.030 & 0.0341 \\
			1.0 & 17964.063 & 17964.137 & 0.0433 & 0.3 & 17964.097 & 17964.166 & 0.0460 & 0.3 & 17964.047 & 17964.168 & 0.0646 \\
			2.0 & 17964.055 & 17964.136 & 0.0431 & 0.4 & 17964.110 & 17964.183 & 0.0453 & 0.4 & 17964.077 & 17964.288 & 0.1102 \\
			3.0 & 17964.056 & 17964.109 & 0.0421 & 0.5 & 17964.030 & 17964.127 & 0.0525 & 0.5 & 17964.233 & 17964.434 & 0.1376 \\
			5.0 & 17963.950 & 17964.050 & 0.0534 & 0.6 & 17964.074 & 17964.133 & 0.0338 & 0.6 & 17964.459 & 17964.694 & 0.1577 \\
			7.0 & 17963.999 & 17964.055 & 0.0308 & 0.7 & 17963.952 & 17964.049 & 0.0351 & 0.7 & 17964.349 & 17964.738 & 0.3273 \\
			10 & \textbf{17963.854} & \textbf{17963.895} & \textbf{0.0172} & 0.8 & 17963.946 & 17964.011 & 0.0347 & 0.8 & 17964.344 & 17964.811 & 0.2679 \\
			15 & 17964.041 & 17964.113 & 0.0515 & 0.9 & \textbf{17963.804} & \textbf{17963.880} & \textbf{0.0185} & 0.9 & 17964.901 & 17965.647 & 0.6320 \\
			20 & 17964.030 & 17964.329 & 0.0875 & 0.99 & 17963.869 & 17963.907 & 0.0227 & 0.99 & 17964.928 & 17965.826 & 0.5852 \\
			Params & \multicolumn{3}{c|}{13/$r_a$/0.7/0.1} & Params & \multicolumn{3}{c|}{13/10/$p_c$/0.1} & Params & \multicolumn{3}{c}{13/10/0.9/$p_m$} \\
			\hline
		\end{tabular}
	\end{center}
\end{table*}

The simulation results are presented in Table \ref{tbl:param}. The best, the mean and the standard deviation (S.D.) of the results are presented. The simulation results indicate that the best parameter combination is $|pop|$/13/10/0.9/0.1. In addition, it can be concluded that the correlation among the tested parameters for solving ELD is not significant due to the observation that the worst results for each test is comparable.

\begin{figure}
	\includegraphics[width=\linewidth]{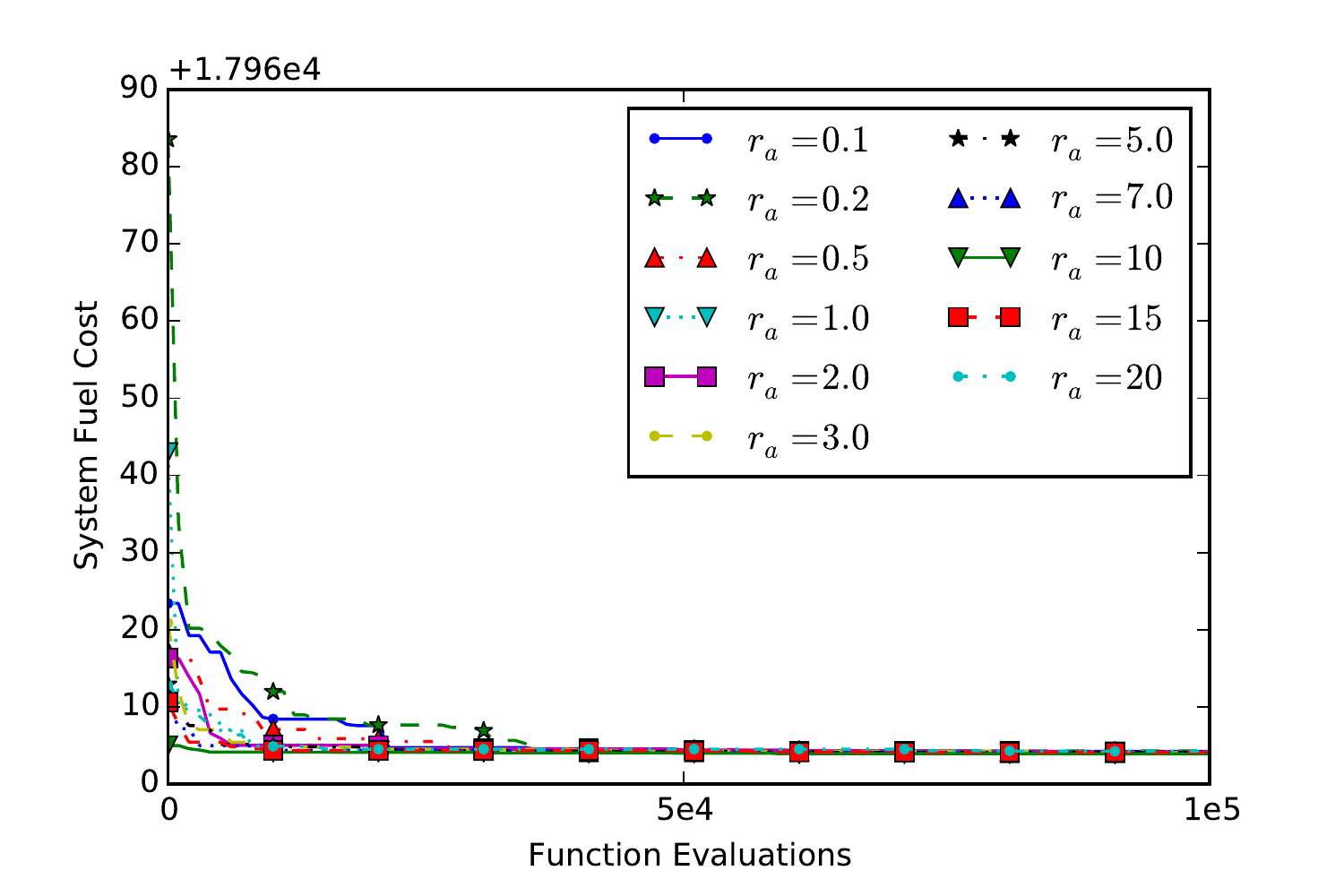}
	\caption{Convergence performance of SSA with different $r_a$ values.}
	\label{fig:conv_ra}
\end{figure}

\begin{figure}
	\includegraphics[width=\linewidth]{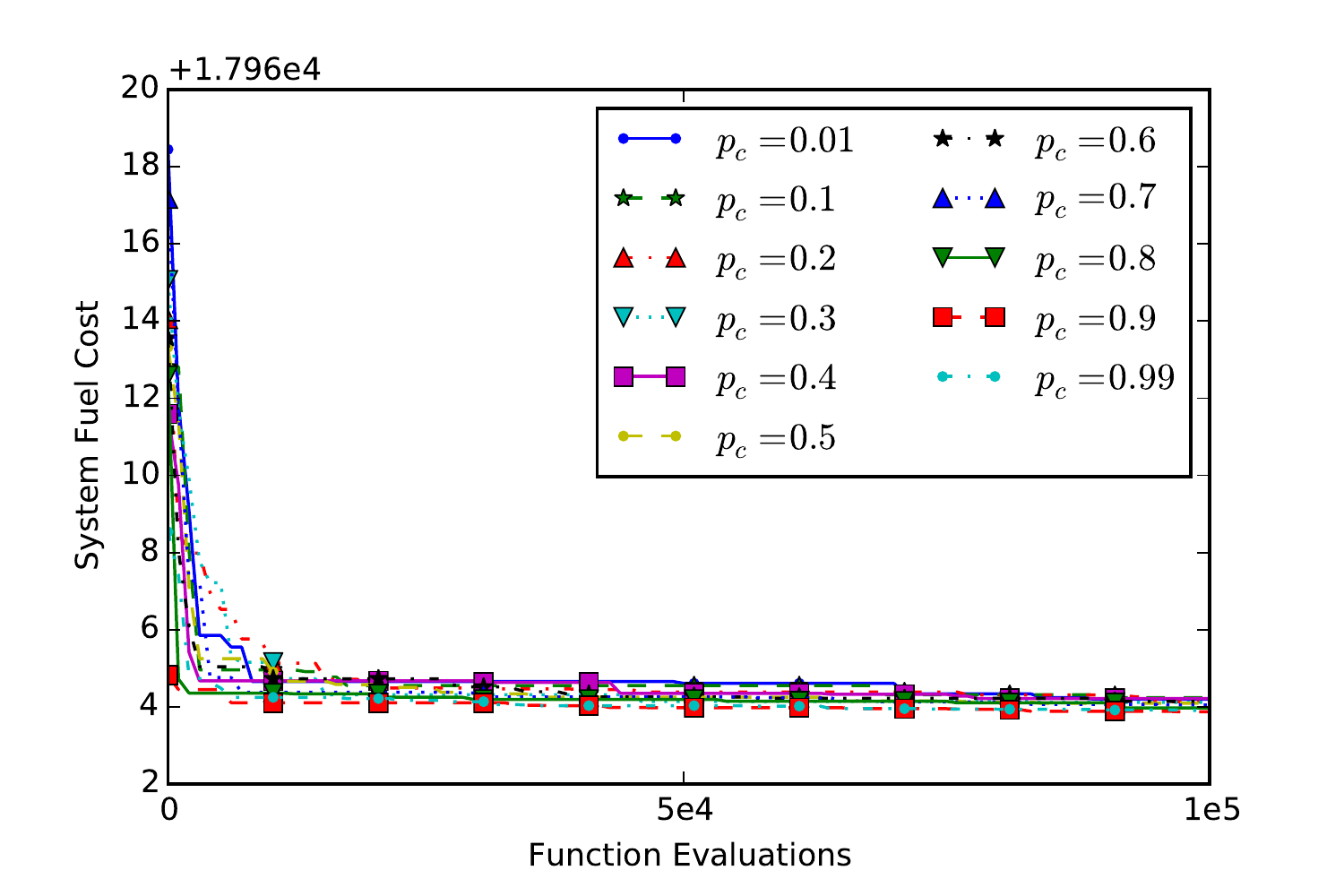}
	\caption{Convergence performance of SSA with different $p_c$ values.}
	\label{fig:conv_pc}
\end{figure}

\begin{figure}
	\includegraphics[width=\linewidth]{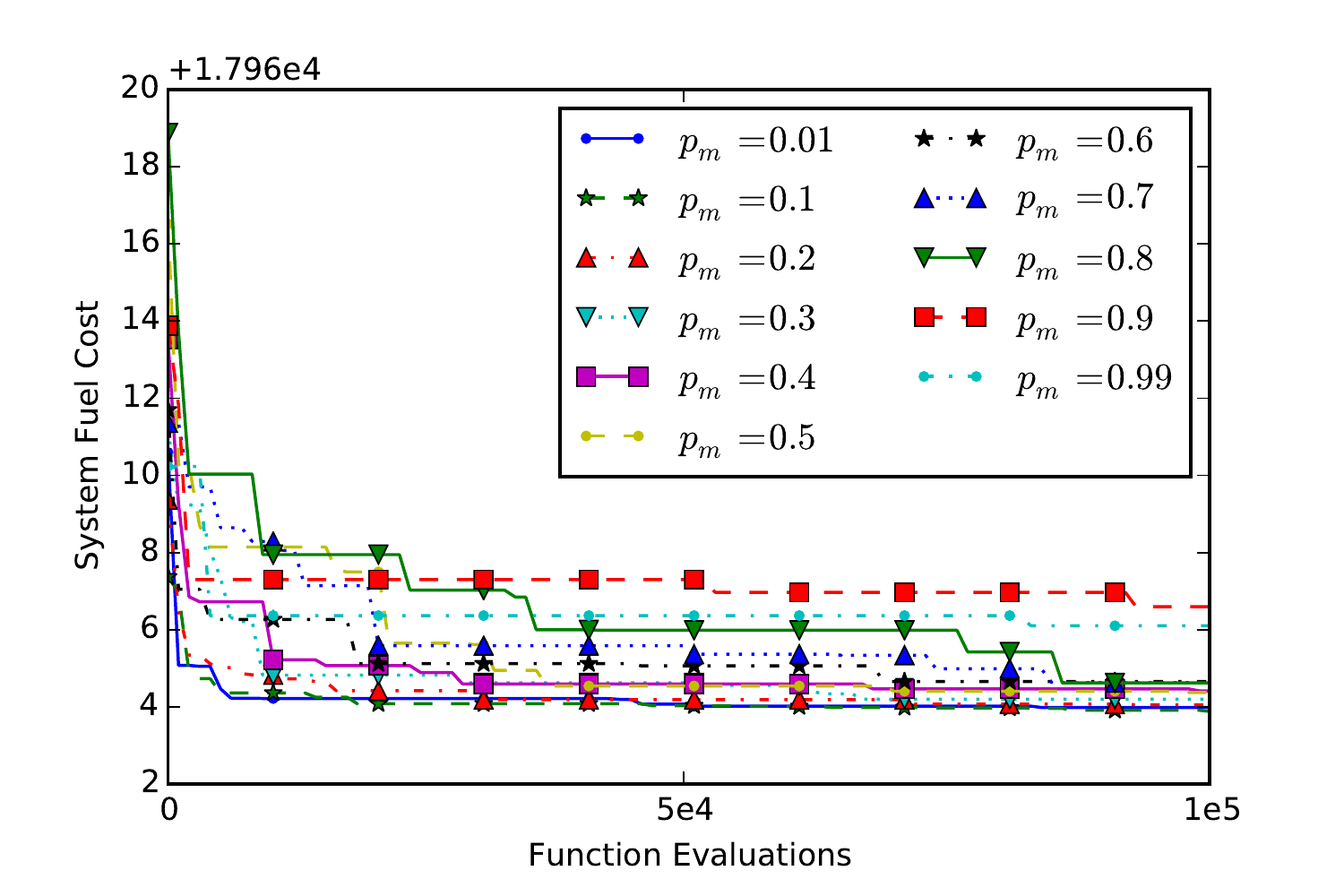}
	\caption{Convergence performance of SSA with different $p_m$ values.}
	\label{fig:conv_pm}
\end{figure}

To better illustrate the convergence performance with respect to different parameter settings, the convergence results of each parameter test are plotted in Figures \ref{fig:conv_ra}, \ref{fig:conv_pc}, \ref{fig:conv_pm}, where the median ones among the 25 runs are presented. The x-axis is the function evaluation counts, and the y-axis is the best-so-far fuel cost. From the results it can be observed that the best performing parameters generally also have the fastest convergence speed.


\section{Conclusion}\label{sec:conclusion}

In this paper we propose a new approach based on the social spider algorithm to solve the economic dispatch problem in power grid operation and control. Although the conventional ELD problem is convex and can be easily solved by mathematical programming methods, the power unit model it employs is not as precise as those considering more practical constraints, e.g., VPE, MFO, and POZ. These characteristics of power units make the optimization problem non-convex, non-differentiable, and non-continuous.

In order to efficiently solve the modern ELD problem, we propose a variant of SSA with new controlling schemes. A chaotic sequence based memory factor is introduced to control the searching pattern, where the previous movement of a spider in SSA is assigned with different degrees of importance with the process of optimization. In addition, we introduce a problem-specific power schedule repairing scheme to fix the infeasible solution generated in the random walk step. This repair scheme takes the power supper/demand and POZ constraints into account, while all other boundary constraints are handled by a boundary absorbing technique.

To evaluate the performance of our proposed SSA-based ELD solver, the approach is applied to solve five different test power systems with various numbers of power units and constraint configurations. The simulation results are compared with a wide range of the state-of-the-art algorithms in solving ELD within the employed test systems. SSA is able to find new best fuel cost solution in four out of the five systems, and can achieve the same solution quality in the remaining one. This result indicates the superiority of SSA in solving ELD with different configurations. In addition, we performed a parameter sensitivity test to develop a best performing combination of SSA parameters. The convergence performance of different parameter values are also presented for comparison. From all simulation results it can be concluded that our proposed SSA-based approach outperforms the existing state-of-the-art algorithms in solving non-convex ELD problems.

\section*{Appendix}

\begin{table}
	\caption{System Coefficients for 13-unit Test System with VPE}
	\label{tbl:13unit_config}
	\scriptsize
	\begin{center}
		\begin{tabular}{crrrrrrr}
			\hline
			Unit($i$) & $P^{\textit{min}}_i$ & $P^{\textit{max}}_i$ & $a_i$ & $b_i$ & $c_i$ & $e_i$ & $f_i$ \\
			\hline
			1 & 0 & 680 & 550 & 8.1 & 0.00028 & 300 & 0.035 \\
			2 & 0 & 360 & 309 & 8.1 & 0.00056 & 200 & 0.042 \\
			3 & 0 & 360 & 307 & 8.1 & 0.00056 & 200 & 0.042 \\
			4 & 60 & 180 & 240 & 7.74 & 0.00324 & 150 & 0.063 \\
			5 & 60 & 180 & 240 & 7.74 & 0.00324 & 150 & 0.063 \\
			6 & 60 & 180 & 240 & 7.74 & 0.00324 & 150 & 0.063 \\
			7 & 60 & 180 & 240 & 7.74 & 0.00324 & 150 & 0.063 \\
			8 & 60 & 180 & 240 & 7.74 & 0.00324 & 150 & 0.063 \\
			9 & 60 & 180 & 240 & 7.74 & 0.00324 & 150 & 0.063 \\
			10 & 40 & 120 & 126 & 8.6 & 0.00284 & 100 & 0.084 \\
			11 & 40 & 120 & 126 & 8.6 & 0.00284 & 100 & 0.084 \\
			12 & 55 & 120 & 126 & 8.6 & 0.00284 & 100 & 0.084 \\
			13 & 55 & 120 & 126 & 8.6 & 0.00284 & 100 & 0.084 \\
			\hline
		\end{tabular}
	\end{center}
\end{table}

\begin{table*}
	\caption{System Coefficients for 40-unit Test System with VPE}
	\label{tbl:40unit_config}
	\scriptsize
	\begin{center}
		\hspace*{-1.0in}
		\begin{tabular}{crrrrrrr|crrrrrrr}
			\hline
			Unit($i$) & $P^{\textit{min}}_i$ & $P^{\textit{max}}_i$ & $a_i$ & $b_i$ & $c_i$ & $e_i$ & $f_i$ & Unit($i$) & $P^{\textit{min}}_i$ & $P^{\textit{max}}_i$ & $a_i$ & $b_i$ & $c_i$ & $e_i$ & $f_i$ \\
			\hline
			1 & 36 & 114 & 94.705 & 6.73 & 0.0069 & 100 & 0.084 & 21 & 254 & 550 & 785.96 & 6.63 & 0.00298 & 300 & 0.035 \\
			2 & 36 & 114 & 94.705 & 6.73 & 0.0069 & 100 & 0.084 & 22 & 254 & 550 & 785.96 & 6.63 & 0.00298 & 300 & 0.035 \\
			3 & 60 & 120 & 309.54 & 7.07 & 0.02028 & 100 & 0.084 & 23 & 254 & 550 & 794.53 & 6.66 & 0.00284 & 300 & 0.035 \\
			4 & 80 & 190 & 369.03 & 8.18 & 0.00942 & 150 & 0.063 & 24 & 254 & 550 & 794.53 & 6.66 & 0.00284 & 300 & 0.035 \\
			5 & 47 & 97 & 148.89 & 5.35 & 0.01140 & 120 & 0.077 & 25 & 254 & 550 & 801.32 & 7.10 & 0.00277 & 300 & 0.035 \\
			6 & 68 & 140 & 222.33 & 8.05 & 0.01142 & 100 & 0.084 & 26 & 254 & 550 & 801.32 & 7.10 & 0.00277 & 300 & 0.035 \\
			7 & 110 & 300 & 287.71 & 8.03 & 0.00357 & 200 & 0.042 & 27 & 10 & 150 & 1055.1 & 3.33 & 0.52124 & 120 & 0.077 \\
			8 & 135 & 300 & 391.98 & 6.99 & 0.00492 & 200 & 0.042 & 28 & 10 & 150 & 1055.1 & 3.33 & 0.52124 & 120 & 0.077 \\
			9 & 135 & 300 & 455.76 & 6.6 & 0.00573 & 200 & 0.042 & 29 & 10 & 150 & 1055.1 & 3.33 & 0.52124 & 120 & 0.077 \\
			10 & 130 & 300 & 722.82 & 12.9 & 0.00605 & 200 & 0.042 & 30 & 47 & 94 & 148.89 & 5.35 & 0.01140 & 120 & 0.077 \\
			11 & 94 & 375 & 635.20 & 12.9 & 0.00515 & 200 & 0.042 & 31 & 60 & 190 & 222.92 & 6.43 & 0.00160 & 150 & 0.063 \\
			12 & 94 & 375 & 654.69 & 12.8 & 0.00569 & 200 & 0.042 & 32 & 60 & 190 & 222.92 & 6.43 & 0.00160 & 150 & 0.063 \\
			13 & 125 & 500 & 913.40 & 12.5 & 0.00421 & 300 & 0.035 & 33 & 60 & 190 & 222.92 & 6.43 & 0.00160 & 150 & 0.063 \\
			14 & 125 & 500 & 1760.4 & 8.84 & 0.00752 & 300 & 0.035 & 34 & 90 & 200 & 107.87 & 8.95 & 0.00010 & 200 & 0.042 \\
			15 & 125 & 500 & 1728.3 & 9.15 & 0.00708 & 300 & 0.035 & 35 & 90 & 200 & 116.58 & 8.62 & 0.00010 & 200 & 0.042 \\
			16 & 125 & 500 & 1728.3 & 9.15 & 0.00708 & 300 & 0.035 & 36 & 90 & 200 & 116.58 & 8.62 & 0.00010 & 200 & 0.042 \\
			17 & 220 & 500 & 647.85 & 7.97 & 0.00313 & 300 & 0.035 & 37 & 25 & 110 & 307.45 & 5.88 & 0.01610 & 80 & 0.098 \\
			18 & 220 & 500 & 649.69 & 7.95 & 0.00313 & 300 & 0.035 & 38 & 25 & 110 & 307.45 & 5.88 & 0.01610 & 80 & 0.098 \\
			19 & 242 & 550 & 647.83 & 7.97 & 0.00313 & 300 & 0.035 & 39 & 25 & 110 & 307.45 & 5.88 & 0.01610 & 80 & 0.098 \\
			20 & 242 & 550 & 647.81 & 7.97 & 0.00313 & 300 & 0.035 & 40 & 242 & 550 & 647.83 & 7.97 & 0.00313 & 300 & 0.035 \\
			\hline
		\end{tabular}
	\end{center}
\end{table*}
\begin{table*}
	\caption{System Coefficients for 10-unit Test System with VPE and MFO}
	\label{tbl:10unit_config}
	\scriptsize
	\begin{center}
		\begin{tabular}{ccrrrrrrr}
			\hline
			Unit($i$) & Fuel($g$) & $P^{\textit{min}}_i$ & $P^{\textit{max}}_i$ & $a_i$ & $b_i$ & $c_i$ & $e_i$ & $f_i$ \\
			\hline
			1 & 1 & 100 & 250 & 26.97 & -0.3975 & 0.002176 & 0.02697 & -3.9750 \\
			1 & 2 & 100 & 250 & 21.13 & -0.3059 & 0.001861 & 0.02113 & -3.0590 \\
			2 & 1 & 50 & 230 & 118.4 & -1.2690 & 0.004194 & 0.11840 & -12.690 \\
			2 & 2 & 50 & 230 & 1.865 & -0.0399 & 0.001138 & 0.00187 & -0.3988 \\
			2 & 3 & 50 & 230 & 13.65 & -0.1980 & 0.001620 & 0.01365 & -1.9800 \\
			3 & 1 & 200 & 500 & 39.79 & -0.3116 & 0.001457 & 0.03979 & -3.1160 \\
			3 & 2 & 200 & 500 & -59.14 & 0.4864 & 0.00001176 & -0.05914 & 4.8640 \\
			3 & 3 & 200 & 500 & -2.876 & 0.0339 & 0.0008035 & -0.00288 & 0.3389 \\
			4 & 1 & 99 & 265 & 1.983 & -0.0311 & 0.001049 & 0.00198 & -0.3114 \\
			4 & 2 & 99 & 265 & 52.85 & -0.6348 & 0.002758 & 0.05285 & -6.3480 \\
			4 & 3 & 99 & 265 & 266.8 & -2.3380 & 0.005935 & 0.26680 & -23.380 \\
			5 & 1 & 190 & 490 & 13.92 & -0.0873 & 0.001066 & 0.01392 & -0.8733 \\
			5 & 2 & 190 & 490 & 99.76 & -0.5206 & 0.001597 & 0.09976 & -5.2060 \\
			5 & 3 & 190 & 490 & -53.99 & 0.4462 & 0.0001498 & -0.05399 & 4.4620 \\
			6 & 1 & 85 & 265 & 52.15 & -0.6348 & 0.002758 & 0.05285 & -6.3480 \\
			6 & 2 & 85 & 265 & 1.983 & -0.0311 & 0.001049 & 0.00198 & -0.3114 \\
			6 & 3 & 85 & 265 & 266.6 & -2.3380 & 0.005935 & 0.26680 & -23.380 \\
			7 & 1 & 200 & 500 & 18.93 & -0.1325 & 0.001107 & 0.01893 & -1.3250 \\
			7 & 2 & 200 & 500 & 43.77 & -0.2267 & 0.001165 & 0.04377 & -2.2670 \\
			7 & 3 & 200 & 500 & 43.35 & 0.3559 & 0.0002454 & -0.04335 & 3.5590 \\
			8 & 1 & 99 & 265 & 1.983 & -0.0311 & 0.001049 & 0.00198 & -0.3114 \\
			8 & 2 & 99 & 265 & 52.85 & -0.6348 & 0.002758 & 0.05285 & -6.3480 \\
			8 & 3 & 99 & 265 & 266.8 & -2.3380 & 0.005935 & 0.26680 & -23.380 \\
			9 & 1 & 130 & 440 & 88.53 & -0.5675 & 0.001554 & 0.08853 & -5.6750 \\
			9 & 2 & 130 & 440 & 15.32 & -0.0451 & 0.007033 & 0.01423 & -0.1817 \\
			9 & 3 & 130 & 440 & 14.23 & -0.0182 & 0.0006121 & 0.01423 & -0.1817 \\
			10 & 1 & 200 & 490 & 13.97 & -0.0994 & 0.001102 & 0.01397 & -0.9938 \\
			10 & 2 & 200 & 490 & -61.13 & 0.5084 & 0.00004164 & -0.06113 & 5.0840 \\
			10 & 3 & 200 & 490 & 46.71 & -0.2024 & 0.001137 & 0.04671 & -2.0240 \\
			\hline
		\end{tabular}
	\end{center}
\end{table*}
\begin{table*}
	\caption{System Coefficients for 6-unit Test System with POZ and Line Loss}
	\label{tbl:6unit_config}
	\scriptsize
	\begin{center}
		\begin{tabular}{crrrrrrrrc}
			\hline
			Unit($i$) & $P^{\textit{min}}_i$ & $P^{\textit{max}}_i$ & $a_i$ & $b_i$ & $c_i$ & $P^{\textit{UR}}$ & $P^{\textit{DR}}$ & $P^{\textit{prev}}_i$ & POZs\\
			\hline
			1 & 100 & 500 & 240 & 7.0 & 0.0070 & 80 & 120 & 440 & $[210,240],[350,380]$\\
			2 & 50 & 200 & 200 & 10.0 & 0.0095 & 50 & 90 & 170 & $[90,110],[140,160]$ \\
			3 & 80 & 300 & 220 & 8.5 & 0.0090 & 65 & 100 & 200 & $[150,170],[210,240]$ \\
			4 & 50 & 150 & 200 & 11.0 & 0.0090 & 50 & 90 & 150 & $[80,90],[110,120]$ \\
			5 & 50 & 200 & 220 & 10.5 & 0.0080 & 50 & 90 & 190 & $[90,110],[140,150]$\\
			6 & 50 & 120 & 190 & 12.0 & 0.0075 & 50 & 90 & 110 & $[75,85],[100,105]$ \\
			\hline
		\end{tabular}
	\end{center}
\end{table*}

\begin{table*}
	\caption{System Coefficients for 15-unit Test System with POZ and Line Loss}
	\label{tbl:15unit_config}
	\scriptsize
	\begin{center}
		\hspace*{-0.5in}
		\begin{tabular}{crrrrrrrrc}
			\hline
			Unit($i$) & $P^{\textit{min}}_i$ & $P^{\textit{max}}_i$ & $a_i$ & $b_i$ & $c_i$ & $P^{\textit{UR}}$ & $P^{\textit{DR}}$ & $P^{\textit{prev}}_i$ & POZs \\
			\hline
			1 & 150 & 455 & 671 & 10.1 & 0.000299 & 80 & 120 & 400 & \\
			2 & 150 & 455 & 574 & 10.2 & 0.000183 & 80 & 120 & 300 & $[185,225],[305,335],[420,450]$ \\
			3 & 20 & 130 & 374 & 8.80 & 0.001126 & 130 & 130 & 105 & \\
			4 & 20 & 130 & 374 & 8.80 & 0.001126 & 130 & 130 & 100 & \\
			5 & 150 & 470 & 461 & 10.4 & 0.000205 & 80 & 120 & 90 & $[180,200],[305,335],[390,420]$ \\
			6 & 135 & 460 & 630 & 10.1 & 0.000301 & 80 & 120 & 400 & $[230,255],[365,395],[430,455]$ \\
			7 & 135 & 465 & 548 & 9.80 & 0.000364 & 80 & 120 & 350 & \\
			8 & 60 & 300 & 227 & 11.2 & 0.000338 & 65 & 100 & 95 & \\
			9 & 25 & 162 & 173 & 11.2 & 0.000807 & 60 & 100 & 105 & \\
			10 & 25 & 160 & 175 & 10.7 & 0.001203 & 60 & 100 & 110 & \\
			11 & 20 & 80 & 186 & 10.2 & 0.003586 & 80 & 80 & 60 & \\
			12 & 20 & 80 & 230 & 9.90 & 0.005513 & 80 & 80 & 40 & $[30,40],[55,65]$ \\
			13 & 25 & 85 & 225 & 13.1 & 0.000371 & 80 & 80 & 30 & \\
			14 & 15 & 55 & 309 & 12.1 & 0.001929 & 55 & 55 & 20 & \\
			15 & 15 & 55 & 323 & 12.4 & 0.004447 & 55 & 55 & 20 & \\
			\hline
		\end{tabular}
	\end{center}
\end{table*}

The system coefficients and configurations are presented in Tables \ref{tbl:13unit_config}, \ref{tbl:40unit_config}, \ref{tbl:10unit_config}, \ref{tbl:6unit_config}, and \ref{tbl:15unit_config}.

\section*{References}

\bibliographystyle{elsarticle-num}
\bibliography{IEEEabrv,../../../bib/publications}

\end{document}